%% file: main.tex
\documentclass[conference]{IEEEtran}

\ifCLASSINFOpdf
  \usepackage[pdftex]{graphicx}
\else
\fi

\usepackage{amsmath}
\usepackage{gensymb}
\usepackage{url}
\usepackage{xcolor}
\usepackage[bottom]{footmisc}
\definecolor{red}{RGB}{255, 0, 0}
\usepackage{multirow}
\usepackage{soul}
\usepackage{siunitx}

\DeclareMathOperator*{\argmax}{arg\,max}

\newcommand\blfootnote[1]{%
  \begingroup
  \renewcommand\thefootnote{}\footnote{#1}%
  \addtocounter{footnote}{-1}%
  \endgroup
}

\begin{document}

\title{Towards a Standardized Reinforcement Learning Framework for AAM Contingency Management}

\author{\IEEEauthorblockN{Luis E. Alvarez}
\IEEEauthorblockA{Surveillance Systems\\
MIT Lincoln Laboratory\\
Lexington, MA USA\\
luis.alvarez@ll.mit.edu}
\and
\IEEEauthorblockN{Marc Brittain}
\IEEEauthorblockA{Surveillance Systems\\
MIT Lincoln Laboratory\\
Lexington, MA USA\\
marc.brittain@ll.mit.edu}
\and
\IEEEauthorblockN{Kara Breeden}
\IEEEauthorblockA{Surveillance Systems\\
MIT Lincoln Laboratory\\
Lexington, MA USA\\
kara.breeden@ll.mit.edu}}
\maketitle

\input{abstract.tex}

\IEEEpeerreviewmaketitle

\blfootnote{DISTRIBUTION STATEMENT A. Approved for public release. Distribution is unlimited. This material is based upon work supported by the National Aeronautics and Space Administration under Air Force Contract No. FA8702-15-D-0001. Any opinions, findings, conclusions or recommendations expressed in this material are those of the author(s) and do not necessarily reflect the views of the National Aeronautics and Space Administration. \copyright~2023 Massachusetts Institute of Technology. Delivered to the U.S. Government with Unlimited Rights, as defined in DFARS Part 252.227-7013 or 7014 (Feb 2014). Notwithstanding any copyright notice, U.S. Government rights in this work are defined by DFARS 252.227-7013 or DFARS 252.227-7014 as detailed above. Use of this work other than as specifically authorized by the U.S. Government may violate any copyrights that exist in this work.

}

\section{Introduction}
Advanced Air Mobility (AAM) is a rapidly evolving field of aviation that has the potential to revolutionize air transportation. AAM encompasses transportation of people and cargo (e.g., passenger shuttling and package delivery) between local, regional, and urban locations with a diverse set of vehicle types which includes small uncrewed aerial systems (sUAS) and larger occupied vehicles such as autonomous conventional takeoff and landing (CTOL) and vertical takeoff and landing (VTOL) aircraft. Due in part to the envisioned high density of operations, AAM requires increasingly autonomous operations to enable operational efficiency and reduce risk in the airspace.

Phases of AAM operations are described by the FAA and NASA as evolving in UAM Maturity Levels (UML) that range from 1 to 6 with increasing levels of expected densities and complexities of roles and responsibilities~\cite{faa_conops,nasa_conops,faa_conopsv2}. For example, UML 1 envisions a low-density human piloted airspace environment, whereas UML 4+ envisions 100s of simultaneous operations over a local region with a mix of human and autonomously piloted aircraft. UML 4+ expects high density, dynamic, and autonomous operations enabled by artificial intelligence (AI) based decision-making algorithms. Therefore, it is important to understand the trade-offs of various AI algorithms by comparing their performance across a range of safety-critical situations.

Contingency management (CM) is a capability within a larger umbrella referred to by NASA and the National Academies as In-Time Aviation Safety Management Systems (IASMS)~\cite{ellis2021concept, NAP25646}. The scope of CM includes (a) planning a set contingencies prior and\slash or during flight, (b) ability to decide which contingency to execute based on available information (e.g., aircraft or system states, hazard level or likelihood) and (c) execution of contingency. In particular, IASMS CM capability has focused on path planning techniques that have an online system monitor various hazards, namely collisions (airborne or ground), loss of power, loss of control, and loss of flight capability~\cite{ellis2021concept}.

To reduce the probability of entering an unsafe state, the majority of these online systems use path planning heuristic algorithms. Predominant algorithms explored are Monte Carlo approaches of path planning where the algorithm finds an optimal path within a time horizon or explores a set of alternate trajectories given the aircraft's dynamics constraints~\cite{9594498, HAGHIGHI2022108453, 8594225}. Example path planning approaches utilize variations of Rapidly-Exploring Random Tree (RRT) and A* to generate an optimized path based on distance or other environmental metrics (e.g., surrounding air traffic). 
Parameter training for these approaches leverage techniques from machine learning such as bagging methods (e.g., bootstrap sampling) with uncertainties represented by random distributions or pre-trained Bayesian belief networks~\cite{doi:10.2514/6.2022-3459, 6631133, https://doi.org/10.1002/rob.21641}. A drawback of these techniques is the requirement for subject matter experts to create and bound the heuristics used by the model.

Reinforcement Learning (RL), a subset of machine learning, alleviates the drawbacks of the above approaches by learning a policy (akin to heuristics) through repeated interaction with an environment to achieve a specified objective. Recently, RL has shown promise in many challenging games such as Go, Atari, Warcraft, and, most recently, Starcraft II with beyond human-level performance~\cite{silver2016mastering, mnih2015human, vinyals2019grandmaster}. In addition, RL has been applied to many real-world air transportation problems such as conflict detection and resolution~\cite{doi:10.2514/6.2021-1952,PHAM2022103463,doi:10.2514/1.I010807}, autonomous separation assurance~\cite{doi:10.2514/1.I010973, 9564466, 9717997}, and collision avoidance~\cite{li2019optimizing}. Furthermore, the Airborne Collision Avoidance System X (ACAS X) family of systems proved RL agents can improve safety and reduce nuisance alerts in real-world operation compared to the heuristic-based systems, in particular the Traffic Alert and Collision Avoidance System (TCAS) that is currently mandated on all commercial aircraft~\cite{Kochenderfer2013,owen2019acas,alvarez2019acas,doi:10.2514/6.2022-3824}. These notable advancements demonstrate the capability of computational learning algorithms to potentially augment and facilitate human tasks in real-world environments.


In this work, a framework is proposed to enable the development and evaluation of RL-based contingency management designs, particularly those that constrain contingency actions to a limited and deterministic set of highly assured maneuvers. This is a principle of the run-time assurance concept used for safety-critical systems. A set of metrics are introduced to evaluate the nominal (unequipped) performance of the framework, providing a baseline for RL research.

\section{Background}


\label{background}


RL is one type of sequential decision making where the objective is to learn a policy in a given environment. RL requires the environment to be formulated as a Markov Decision Process (MDP), a mathematical framework for modeling decision making processes with stochastic transitions. An MDP can be defined by the five-tuple $(S,A,R,T,\gamma)$, where $S$ is the state space, $A$ is the action set, $R$ is the reward function $R(s,a)$ providing the reward for taking action $a$ from state $s$, and $T$ is the probability transition function, $T(s,a,s')$, that produces the probability of transition from state $s$ to the next state $s'$ when taking action $a$. In RL, the transition matrix $T$ is often unknown. The discount factor $\gamma$ determines how far in the future to look for rewards.  As $\gamma\rightarrow$~0, immediate rewards are emphasized, whereas, when $\gamma\rightarrow$~1, future rewards are prioritized.

From this formulation, the RL agent is able to derive an optimal policy in the environment by maximizing a cumulative reward function. Let $\pi$ represent some policy and $\tau$ represent the total time for a given environment, then the optimal policy can be defined as follows:
\begin{equation}
    \pi^{*} = \argmax_{\pi}E[\sum_{t=0}^{\tau}(r(s_{t}, a_{t})|\pi)].
\end{equation}
By designing the reward function to reflect the objective in the environment, the optimal solution can be obtained by maximizing the total reward.

In environments with a low dimensional state-action space, or environments with discrete state-actions values, the policy $\pi$ can be represented as a look-up table and solved using dynamic programming approaches such as Q-learning~\cite{watkins1992q}. However, in many real-world environments the state-action space can not be represented by discrete values, or the dimensionality of the state-action space is too large to construct a look-up table. Deep reinforcement learning (DRL) alleviates the aforementioned issues by introducing a neural network to represent the policy $\pi$. Various approaches exist for training DRL agents including value-based methods~\cite{mnih2015human, c51, HER},  and policy-based optimization~\cite{trpo, schulman2017proximal, a3c}.

Large scale simulation capabilities are essential for RL algorithms as they explore the transitions between states as a consequence of their actions within the modeled environment. Recently, MIT Lincoln Laboratory introduced the AI testbed for advanced air mobility, or AAM-Gym to provide a representative fast-time simulation environment for research and development of decision making tools for AAM concepts of operation~\cite{aamGymMarc}.
By leveraging AAM-Gym, new RL use-cases, such as contingency management, can be developed to explore the trade-offs of AI/RL-based decision-making functions with existing approaches. The proposed RL contingency management framework utilizes AAM-Gym~\cite{aamGymMarc} with BlueSky~\cite{BlueSky} as the back-end simulation capability to explore the impact of RL algorithm choice, hazards (e.g., wind fields, rate of loss of navigation, battery discharge prediction failures), and aircraft parameters on the overall training and effectiveness of agents. By utilizing AAM-Gym, the state space of each agent can make use of all the aircraft and environment information if desired, as seen in Tables \ref{aircraft_params} and \ref{environment_params}.

In order to provide representative real world scenarios to AAM-Gym, we leverage the Lincoln Laboratory UAMToolKit framework; an event-driven simulation environment that can generate spatio-temporal traffic operations with a diverse set of flight routes~\cite{doi:10.2514/6.2021-2381}. When modeling Urban Air Mobility (UAM) operations, time variant cycles of demand across the New York City metropolitan area are provided by estimating the demand as a percentage of the expected replacement by AAM of for-hire-vehicle (FHV) operations (e.g. Uber, Lyft, taxi, limousine services). The event driven model outputs a complete history of aircraft movements and metadata, which includes departure times, cruising speeds, arrival times, flight route (i.e., flight level and waypoints), origin-destination pair, and aircraft type. UAMToolKit can be queried by AAM-Gym allowing for vast amounts of unique and diverse traffic patterns for the RL agents to train and evaluate against.

\begin{table}[tb]
\caption{Aircraft Information Available for Agent Training.}
\centering
\label{aircraft_params}
\begin{tabular}{lll}
Parameter                    & Range &  \\ \hline
\multicolumn{1}{l|}{Heading} &  0 - 360\degree    &  \\
\multicolumn{1}{l|}{Altitude} &  0 - 5,000 ft    &  \\
\multicolumn{1}{l|}{Aircraft horizontal speed} &  0 - 120 knots    &  \\
\multicolumn{1}{l|}{Horizontal acceleration} &  0.1 - 0.5$g$     &  \\
\multicolumn{1}{l|}{Flight path distance to destination} &  -     &  \\
\multicolumn{1}{l|}{Energy level} &  0 - 350 KWh    &  \\
\multicolumn{1}{l|}{Flight route waypoints} & 0 - complete route     &  \\
\hline
\end{tabular}
\end{table}

\begin{table}[tb]
\caption{Environment Information Available for Agent Training.}
\centering
\label{environment_params}
\begin{tabular}{ll}
Parameter                    & Notes   \\ \hline
\multicolumn{1}{l|}{Wind} &  North and east magnitudes      \\
\multicolumn{1}{l|}{Vertiport locations} &  Latitude and longitude      \\
\multicolumn{1}{l|}{Approximate population density} &  From NYC FHV database    \\
\multicolumn{1}{l|}{Probability of loss of navigation ($P_{nav}$)} &  Guassian noise per aircraft \\
\multicolumn{1}{l|}{Aircraft in vicinity} &  All fields available per aircraft     \\
\hline
\end{tabular}
\end{table}

\section{Approach}
In this section, the expansions of AAM-Gym for the RL contingency management use-case are described as well as the MDP formulation.

\subsection{Traffic Generation}
When modeling UAM operations, UAMToolKit requires information about the fleet size, aircraft capacity of each vertiport,  vehicle passenger capacity, cruising speeds, the corridor network for operation, distributions of turn-around time, and wind data. 
The focus of this research is on contingency management during flight and not scheduling impacts due to turn-around and wind data. Therefore, operations are modeled with zero wind field and a one minute of turn-around time, the minimum simulation step.
Operations occur between eight altitude lanes evenly spaced from 1,000--5,000 ft utilizing the existing helicopter network as depicted in~\cite{doi:10.2514/6.2021-2381}. Passenger capacity is maintained at four passengers, and 50--500 aircraft are attempting to meet 5\% of the FHV demand, with landing pad spaces in each vertiport maintained so all aircraft have a landing pad at the beginning of the simulation and an additional landing pad for emergency. For example, in the case of the 500 fleet size scenario, each of the 29 vertiports selected in~\cite{doi:10.2514/6.2021-2381} would have 18 landing pads. 

\subsection{Algorithms}
AAM-Gym allows researchers to develop, train, and evaluate RL algorithms that adhere to the OpenAI Gym~\cite{https://doi.org/10.48550/arxiv.1606.01540} protocol, allowing for rapid prototyping of new and existing algorithms. AAM-Gym has support for both general RL based frameworks such as Ray's RL-Lib~\cite{liang2018rllib} and air transportation based RL algorithms such as the Deep Distributed Multi-Agent Variable with Attention (D2MAV-A)~\cite{doi:10.2514/6.2021-1952}.

\subsection{Hazard Modeling}
The hazards initially modeled by the RL framework include: uncertainty in battery consumption, max charge capacity due to number of charge cycles, wind information, and navigation capability. Other hazards will be modeled in the future.

As part of the RL contingency management use-case, two energy consumption rate models were developed. The energy consumption rate is modeled within the RL framework as either a linear consumption model or a mixture of a BADA 3 and Bada H~\cite{nuic2010bada, mouillet2018bada} based energy consumption model during cruising phases with a linear model at speeds below 20 knots.

The linear energy consumption model is represented by (\ref{eq:consumption_rate}) where $\alpha$ is the constant energy consumption rate per time step, $\Delta{t}$ is the time step between simulation iteration, $C$ is the total number of charge cycles an aircraft has experienced, and $\beta$ is the number of charge cycles when uncertainty in consumption begins to occur. The uncertainty in consumption rate after $\beta$ charge cycles is modeled as a truncated Gaussian distribution, $N$, of range $[0,1]$ for values above $\phi$.
\begin{equation}
    \label{eq:consumption_rate}
    \Delta{E} =
    \begin{cases}
        \alpha \cdot \Delta{t} & \text{if $C\leq\beta$}  \\
        \alpha \cdot \Delta{t} +\text{$N$} & \text{if $C>\beta$}  \\
    \end{cases}.
\end{equation}

\begin{figure}[t]
\centering
\includegraphics[width=0.45\textwidth]{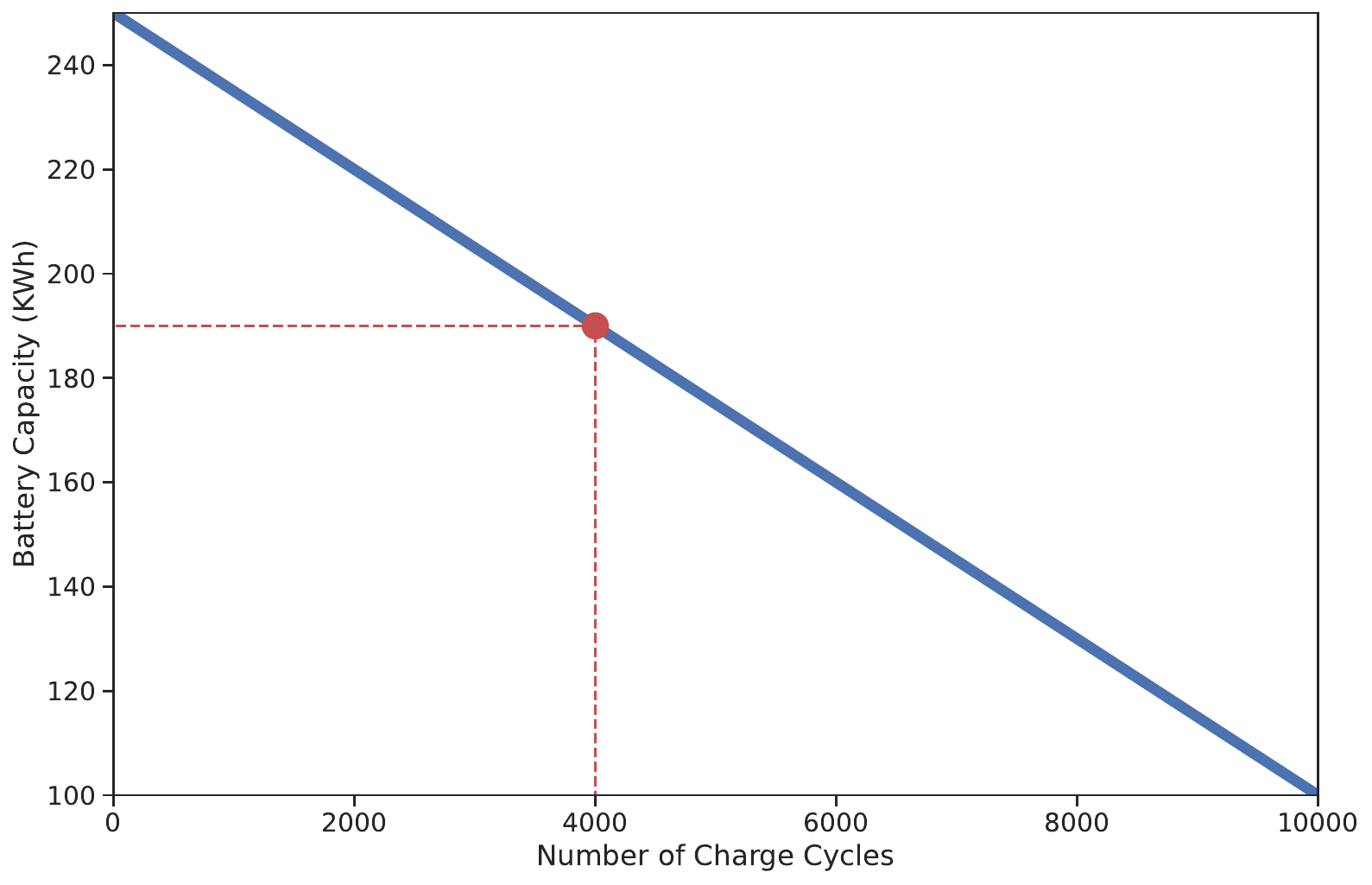}
\caption{Example of randomly sampled energy assigned to aircraft at creation (red circle) from a linear function of charge cycles and energy level when $E_{max}=250$ KWh, $C_{min}=0$ cycles, $E_{min}=100$ KWh and $C_{max}=10,000$ cycles.}
\label{fig:charge_vs_cycles}
\end{figure}

 When each aircraft is created, it is randomly assign a value for the number of charge cycles, $C$, by a uniform distribution of a given range. The model then uses the defined minimum energy level ($E_{min})$, maximum energy level ($E_{max}$), minimum charge cycles ($C_{min}$), and maximum charge cycles ($C_{max}$) to create a linear fit and assign the maximum energy level possible given the number of cycles the aircraft has experienced (example shown in Fig. \ref{fig:charge_vs_cycles}).
 

The wind fields can be provided as constant wind fields, input data from Weather Research Forecast (WRF) model~\cite{JamesTim2023}, or as multivariate distributions. When providing discrete locations for wind field information, the BlueSky simulation environment uses multi-linear interpolation to provide the wind vector at specified locations.

The probability of navigation loss, $P_{nav}$, can also be modeled as a multivariate distribution providing the probability of losing navigation at specified latitude and longitudes or as a discrete event occurring with a set probability.

\subsection{Reward Model and Terminal States}

As previously presented, the RL framework uses an MDP to construct the contingency management problem where the researchers define the state space, actions, transitions, and rewards for each agent. In this framework the transition model is unknown as its dictated by the simulation environment and the hazard parameter chosen. Therefore, it is not explicitly defined by the researchers. The action space of the MDP evaluated is shown in Table \ref{action_sets}. The available actions are defined as follows:

\begin{enumerate}
    \item Heading change - The aircraft turns left or right by a pre-defined magnitude (e.g., $\pm$5$\degree$, 0$\degree$) and continues on this heading until a new action is chosen.
    \item Land now - The aircraft chooses to execute a controlled vertical decent from present position.
    \item No alert - The aircraft takes no action and continues on its current heading or active flight plan.
    \item Use assigned flight route - The aircraft continues on its planned flight route or returns to the nearest waypoint, if prior actions caused a deviation, and continues along flight route.
\end{enumerate}


\begin{table}[tb]
\caption{Actions available for agents}
\centering
\label{action_sets}
\begin{tabular}{lll}
Description & Value\\ \hline
Heading change &  -5\degree,0,-5\degree       \\
Land now &  -    \\
No alert &  -     \\
Use assigned flight route & -   \\
\hline
\end{tabular}
\end{table}

The reward function utilized by the MDP is defined by 
\begin{equation}
\label{reward_func}
R(s_t,h_t,a_{t}) = R(s_t) + R(h_t) + R(a_{t}) - \Omega,
\end{equation}
where $R(s_{t})$, $R(h_t)$, and $R(a_{t})$ are defined by~(\ref{eq:r_st}),~(\ref{eq:r_ht}),~(\ref{eq:r_at}) and $\Omega$ is a step penalty.
$R(s_t)$ is a state dependant reward function based on the terminal states of the aircraft as defined by (\ref{eq:r_st}). The RL framework allows for three types of terminal states to be defined as total loss of energy, loss of navigation, and when an aircraft reaches an altitude of zero feet, each respectively adding a penalty $\delta_{energy}$, $\delta_{navigation}$, and $\delta_{range\_to\_destination}$. 
\begin{equation}
\label{eq:r_st}
R(s_t) = 
      \delta_{energy}+\delta_{navigation}+\delta_{range\_to\_destination}
\end{equation}
A multivariate Gaussian distribution is used to define the rewards of landing at destination and alternative vertiports. It is defined as 
\begin{equation}
\label{eq:r_ht}
R(h_t) = \sum_{i=0}^n \delta_{vertiport} \cdot e^{\frac{(X-\mu_{xi})^2+(Y-\mu_{yi})^2}{-2\sigma^2}}
\end{equation}
where $X$ and $Y$ are the latitude and longitude of the aircraft under control. The latitude and longitude of each vertiport are defined by $\mu_x$ and $\mu_y$ respectively. The standard deviation $\sigma$ is set to 0.0005\degree. The variable $\delta_i$ is defined by
\begin{multline}
\delta_{vertiport} = \\
    \begin{cases}
        \delta_{vertiport\_destination} & \text{if vertiport = destination}\\ 
        \delta_{vertiport\_other} & \text{if vertiport $\ne$ destination}\\ 
    \end{cases}.
\end{multline}
\newline
Last, $R(a_{t})$ is defined by
\begin{multline}
\label{eq:r_at}
R(a_{t}) = \\
    \begin{cases}
      \delta_{land} + \delta_{action\_penalty} & \text{if $a_{t} =$ ``Land now''}\\
      \delta_{action\_penalty} & \text{if $a_{t}\neq$ ``No alert''}
    \end{cases},
\end{multline}
where $\delta_{action\_penalty}$ is a penalty applied when any action is chosen other than ``No alert'' and $\delta_{land}$ is the penalty for choosing to ``Land now''.

\subsection{Metrics}
The RL framework extends the capabilities of AAM-Gym and allows users to define performance metrics to help assess the CM algorithm within a Hydra configuration file~\cite{Yadan2019Hydra} leveraging the aircraft and environment information, as defined in Table \ref{aircraft_params} and \ref{environment_params}.
In addition, the following information was recorded: spatial distance from corridor, number of aircraft that reached their destination, energy level of each aircraft, total reward experienced by each aircraft, average wind in scenario, time spent in flight route, and actions taken. 


\section{Numerical Experiment and Results}
Preliminary experiments conducted to date are described in this section with the parameters explored and their resulting impact on the operations. Additional experiments are planned.

\subsection{Parameter Sweep}
The Hydra configuration interface of AAM-Gym allows for multiple experiments to be launched simultaneously to explore the impact of environment decisions and rewards. This experiment explores the impact of modeling an increased consumption rate on the battery model, differences in maximum charge capacity, and activating probability of loss of navigation, $P_{nav}$, per aircraft per simulation step. 
The parameters that were explored and the range of values chosen are shown in Table \ref{tab:param_sweep_env}. For this paper we explore the impact of each of these parameters on the likelihood that an operation will reach its destination or other assigned vertiports as a function of training steps.


\begin{table}[tb]
\caption{Environment Parameters Explored and Values}
\centering
\label{tab:param_sweep_env}
\begin{tabular}{lll}
Parameter & Values Explored  \\ \hline
Maximum initial energy level ($E_{max}$) & 50, 150, 250 KWh \\
Minimum initial energy level ($E_{min}$) & 100 KWh\\
Maximum battery cycles ($C_{max}$)& \num{10000} \\
Minimum battery cycles ($C_{min}$)& 0 \\
Energy consumption uncertainty threshold  ($\beta$)  & \num{3000} \\
Gaussian truncation parameter ($\phi$) & 0.5 \\
Probability of loss of navigation capability ($P_{nav}$) & 0.0, \num{1e-5} \\
\hline
\end{tabular}
\end{table}

\subsection{Results}
\begin{figure}[t]
\centering
\includegraphics[width=0.45\textwidth]{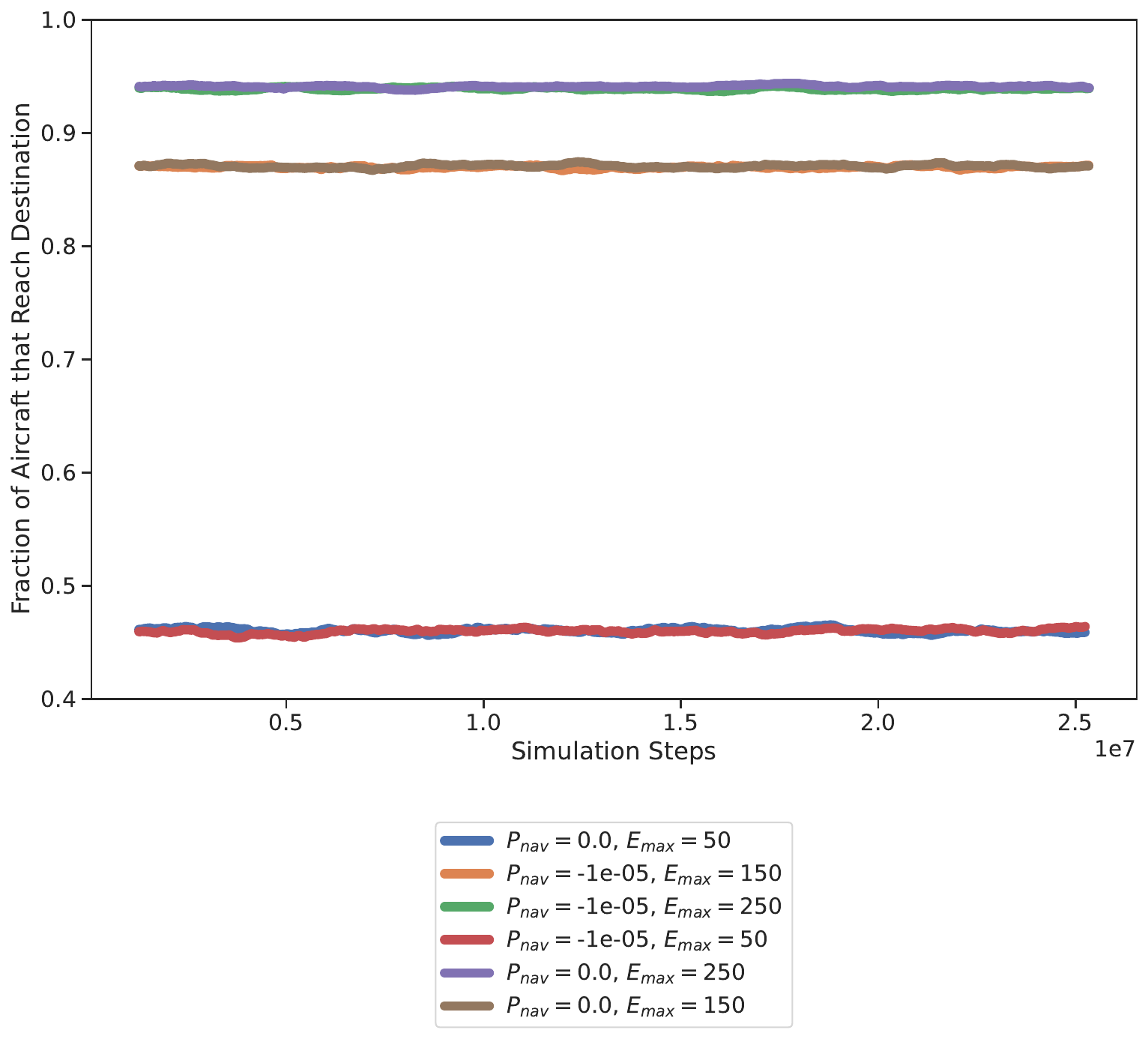}
\caption{Rolling mean of the fraction of aircraft reaching their destination as a function of simulation steps, probability of loss of navigation ($P_{nav}$), and maximum initial energy level ($E_{max}$).}
\label{fig:destination_reached}
\end{figure}

When no RL-based CM agent is in control of the aircraft (i.e., unequipped) the likelihood of reaching their destination is strongly correlated to the maximum initial energy level, $E_{max}$. This trend is evident by the higher fraction of aircraft reaching the destination, $\bar{P}_{dest}$, as seen in Fig. \ref{fig:destination_reached}. Due to all AAM-Gym parallel workers initializing their respective simulations at the same time with a low number of aircraft, we see a spike in fraction of aircraft reaching their destination in Fig. \ref{fig:destination_reached}. As the AAM-Gym parallel workers continue the simulation, the number of aircraft stabilizes, thus creating a steady number of aircraft reaching their destination as a function of $E_{max}$ and $P_{nav}$.


This is expected as operations with lower energy levels are at risk of quickly depleting their energy reserves during flight without reaching any vertiport if an agent does not dictate an action to undertake. 
The unequipped results also show the number of aircraft reaching their final destination decreases when the probability of navigation loss, $P_{nav}$, is above zero, as seen in Table \ref{tab:results_param_sweep_env}. A change in $P_{nav}$ from 0 to \num{1e-5} results in an decrease of aircraft arrivals by 0.12\% to 0.25\% as it increases the likelihood of entering a terminal state.


\begin{table}[tb]
\caption{Simulation Parameters and Metrics of Interest}
\centering
\label{tab:results_param_sweep_env}
\begin{tabular}{cccc}
$P_{nav}$ &		$E_{max}~ \text{(KWh)} $ &		$\max \bar{P}_{dest}$ \\ \hline
 \num{1e-5}      & 50.0   &  0.4639 \\  
 0.0     & 50.0   & 0.4654  \\ 
 \num{1e-5}      & 150.0  & 0.8719 \\ 
 0.0      & 150.0  & 0.8745 \\ %
 \num{1e-5}      & 250.0  &  0.9414 \\ 
 0.0     & 250.0  & 0.9439 \\ 
\hline
\end{tabular}
\end{table}


\section{Conclusion}
In this work, a RL framework was developed to enable standardized research and development in the area of contingency management for AAM. It is shown how the problem is formulated as a Markov Decision Process and through using AAM-Gym and UAMToolkit, a representative simulation environment is established for training and validation of RL agents. Preliminary results are presented to demonstrate the impact of various simulation parameters that induce a contingency event. 

In future work, further analysis of the reward model, MDP environment structure, and algorithm design is required to assess the viability of RL algorithms in contingency management. While baseline results were presented in this study, future work will explore various state-of-the-art RL approaches to understand the viability of learning-based algorithms. After a broad analysis of the performance of these algorithms, a selected set will be tuned through the use of surrogate optimization. In addition, the simulation environment will be further refined with the introduction of crowd source population databases for causality penalties, and use of eVTOL vehicle performance models with improved battery models. 


\section{Acknowledgment}
The authors would like to thank Steve Young of NASA for his guidance and support of this effort. The work was funded by NASA’s System-Wide Safety Project via an Interagency Agreement with the US Air Force (IA 80LARC23TA002, Project 10384).


\bibliographystyle{IEEEtran}
\bibliography{bibliography}

\end{document}

%% file: abstract.tex
\begin{abstract}

Advanced Air Mobility (AAM) is the next generation of air transportation that includes new entrants such as electric vertical takeoff and landing (eVTOL) aircraft, increasingly autonomous flight operations, and small UAS package delivery. With these new vehicles and operational concepts comes a desire to increase densities far beyond what occurs today in and around urban areas, to utilize new battery technology, and to move toward more autonomously-piloted aircraft. To achieve these goals, it becomes essential to introduce new safety management system capabilities that can rapidly assess risk as it evolves across a span of complex hazards and, if necessary, mitigate risk by executing appropriate contingencies via supervised or automated decision-making during flights.
Recently, reinforcement learning has shown promise for real-time decision making across a wide variety of applications including contingency management. In this work, we formulate the contingency management problem as a Markov Decision Process (MDP) and integrate the contingency management MDP into the AAM-Gym simulation framework. This enables rapid prototyping of reinforcement learning algorithms and evaluation of existing systems, thus providing a community benchmark for future algorithm development. We report baseline statistical information for the environment and provide example performance metrics. 

\end{abstract}